\documentclass[a4paper]{article}

\usepackage{INTERSPEECH2021}

\usepackage[export]{adjustbox}

\usepackage{hyperref}

\usepackage{makecell}
\usepackage{multirow}
\usepackage{adjustbox}
\usepackage{tabularx}
\usepackage{subcaption}
\usepackage{caption}
\usepackage{multicol}
\usepackage{graphicx}
\usepackage{float}
\usepackage{color}
\usepackage{amssymb} 
\usepackage{amsmath} 

\DeclareMathOperator*{\argmax}{argmax} 
\DeclareMathOperator*{\softmax}{softmax} 
\DeclareMathOperator*{\repeatnn}{repeat} 
\DeclareMathOperator*{\flatten}{flatten} 

\title{Bi-directional Joint Neural Networks for Intent Classification and Slot Filling}

\name{Soyeon Caren Han*, Siqu Long*\thanks{*co-first author}, Huichun Li, Henry Weld, Josiah Poon}

\address{School of Computer Science, University of Sydney, Australia}
\email{\{Caren.Han, slon6753, huli7926, hwel4188, Josiah.Poon\}@sydney.edu.au}

\begin{document}
\maketitle
\begin{abstract}
Intent classification and slot filling are two critical tasks for natural language understanding. Traditionally the two tasks proceeded independently. However, more recently joint models for intent classification and slot filling have achieved state-of-the-art performance, and have proved that there exists a strong relationship between the two tasks. In this paper, we propose a bi-directional joint model for intent classification and slot filling, which includes a multi-stage hierarchical process via BERT and bi-directional joint natural language understanding mechanisms, including intent2slot and slot2intent, to obtain mutual performance enhancement between intent classification and slot filling. The evaluations show that our model achieves state-of-the-art results on intent classification accuracy, slot filling F1, and significantly improves sentence-level semantic frame accuracy when applied to publicly available benchmark datasets, ATIS (88.6\%) and SNIPS (92.8\%).
\end{abstract}



\section{Introduction}





The introduction of deep learning architectures has shown great success in Natural Language Understanding (NLU), including two main tasks, slot filling and intent classification, which aim to extract the user’s intent and to identify semantic constituents from the natural language utterance. Traditionally the two are processed independently. Slot filling can be treated as a sequence labeling task that annotates an input word sequence [$x_{1}, x_{2}, \ldots, x_{n}$] with a corresponding slot label sequence [$y_{1}^{s}, y_{2}^{s}, \ldots, y_{n}^{s}$], each from a set of slot labels $\mathbb{S}$ \cite{xu2013convolutional}. Intent detection can be considered as a classification problem to map the input sequence to an intent label $y^{I}$ from a set of intent labels $\mathbb{I}$.
Traditional independent models in pipeline usually suffer from error propagation since such models do not consider the mutual relationship between the two tasks. 
This mutual relationship between intent classification and slot filling can be illustrated with the example.
With a word \textit{westbam} being identified as slot \textit{artist}, the utterance query is more likely to be classified as music related, say \textit{PlayMusic}, than from another area, like \textit{BookRestaurant}. Conversely, with the intent recognised as \textit{PlayMusic}, the slot categories are more likely to be related to music (e.g. \textit{artist}, \textit{album}) than other fields, like \textit{restaurant\_type}.  
A large body of work \cite{guo2014joint,hakkani2016multi,zhao2016attentionbased,zhang2016mining,xia2018zero,zhang2019joint,tang2020end} has confirmed that conducting intent classification and slot filling simultaneously achieves a synergistic effect and improve performance over independent models. Earlier joint models typically adopted a single RNN model to fill slots and then detect an intent with a shared objective or, a detected intent informed the filling of slots. Such models achieved decent performance but did not entirely and/or explicitly consider the interrelationship between the intent and slot.

In this paper, we introduce the bi-directional joint model for intent classification and slot filling, which contains a multi-stage hierarchical process using BERT (Bi-directional Encoder Representations from Transformers) \cite{devlin2019bert} and a bi-directional NLU mechanism. The proposed model produces initial intent and slot probabilities from BERT word embeddings, then applies the Bi-directional NLU mechanism to obtain the mutual enhancement between intent classification and slot filling. Our Bi-directional NLU mechanism contains two models: intent2slot for slot filling and slot2intent for intent classification. The intent2slot draws on the intent probability distribution, utilising the semantic information of the whole sequence while labeling each word. The slot2intent takes in the sequence label probability distributions as supplementary information for intent classification. To jointly model intent classification and slot filling, the model is trained on the joint loss of slot labeling and intent classification. 
The contributions are: 1) A bi-directional joint NLU mechanism with intent2slot and slot2intent models achieves full synergistic effects for joint intent classification and slot filling; 2) The proposed model introduces a multi-stage hierarchical process integrating Transformer-based Modelling and a Bi-directional NLU mechanism. 


\section{Related work}



Handling the two sub-tasks in a pipeline has achieved decent results, but suffers from error propagation; a misidentified intent can prompt seemingly matching but incorrect slots in the second step, or vice versa. Models that work on the tasks simultaneously are proposed, many of which implicitly model joint distributions via joint loss back-propagation \cite{guo2014joint, hakkani2016multi, liu2016attention}. However, explicitly capturing the relationships between intents and slots is beneficial. Goo \cite{goo2018slot} used a slot-gate attention-based joint model that enables the learned intent result to assist the slot filling process and performs well in semantic accuracy due to this global consideration. Zhang \cite{zhang2019capsule} proposed a capsule neural network-based joint model that has slot prediction inform intent, and feedback from intent to slot prediction. Chen \cite{chen2019bert} proposed to use the contextual BERT model to jointly learn the two tasks. However, they simply make prediction based on BERT output hidden states without further exploration of the potential relationship between the two tasks. Stack-Propagation-BERT \cite{qin2019stack} uses an intent2slot architecture with BERT encoding and using stack propagation. Stack propagation in multi-task architectures provides a differentiable link from one task to the other rather than performing each one in parallel.

\section{Methodology}


\subsection{Input embedding layer}
The input data to intent detection and slot filling tasks is user utterances in the form of text sentences, which are typically tokenised into sequences of word tokens. The output for slot filling is a sequence of slot labels, the length of which exactly matches the number of tokens in the input. For intent detection the output is a single intent label. The sets of distinct slot labels and intent labels are converted to numerical representations by mapping them to integers. Let $n$ be the maximum input sequence length in the dataset. All input sequences are post-padded to length $n$ with a [PAD] token. Slot label sequences are also padded to length $n$ with a padding label. We use the pre-trained BERT-BASE model for numerical representation of the input sequences. The feature size $N=n+2$ is fed to the model, to include the BERT [CLS] and [SEP] tokens.

\subsection{NLU modelling layer}
BERT gives a contextual, bi-directional representation of input tokens. These representations will be used to compute the initial intent probability and slot probabilities. The BERT model we used is ``bert{\_}uncased{\_}L-12{\_}H-768{\_}A-12''; that is, using uncased tokens, with 12 encoder stacks, final hidden layer of dimension $D_{E}$ = 768, and 12 attention heads. Each stack contains a multi-head attention module and a 2-layer feed forward neural network (FFNN). Within the attention layers there is dropout on attention with probability 0.1. The FFNN contains two dense, linear layers of size 3072 and 768 respectively. A GELU activation function is used for the first layer only. Dropout with probability 0.1 is applied. A residual connection of the attention output is added to the output of the second layer. This addition is normalized and becomes the input for the next encoder stack, and also the final output of the current encoder stack. The output shape is [$N$, $D_{E}$] $\in \mathbb{R}^{N \times D_{E}}$. We add one more dropout layer with dropout rate 0.1 on the encoder output to further mitigate potential overfitting.

\subsection{Bi-directional NLU layer}
On top of the NLU Modelling Layer, we propose the Bi-directional NLU layer that contains two models: intent2slot for slot filling and slot2intent for intent classification. The intent2slot draws on the intent probability distribution, utilising the semantic information of the whole sequence while labeling each slot. The slot2intent takes in the sequence label probability distributions as supplementary information for intent classification. To jointly model intent classification and slot filling, the model is trained on the joint loss of slot labeling and intent classification.

\textbf{Intent2Slot model}
The intent2slot model aims to draw the intent probability by extracting the semantic information of the whole sequence and utilising it to aid detection of a slot label for each word. Let $D_{IP}$  and  $D_{SP}$  be the number of distinct intent labels and slot labels respectively. Let the encoder output of size [$N$,$D_{E}$] be denoted $(h_{1}, h_{2}, \ldots, h_{N})$ where each $h_{i}$ is the $D_{E}=768$ dimension final hidden layer state of each token. Here $h_{1}$ is the encoding of the [CLS] token which is trained to classify the entire sequence. In order to utilise the intent probability $P_{I}$ for slot filling we first take $h_{1}$ and pass it through a dense layer then apply a softmax: 
\begin{equation}
     P_{I}= \softmax(h_{1} \cdot W_{I}^{T} + b_{I})
\end{equation}

This is a dense layer with $D_{IP}$ nodes. Hence $W_{I} \in \mathbb{R}^{D_{IP} \times D_{E}}$ is a matrix of weights and $b_{I} \in \mathbb{R}^{D_{IP}}$ is a vector of biases. $P_{I}\in \mathbb{R}^{D_{IP}}$ is thus a vector of probabilities. We next broadcast $P_{I}$ to the length of sequence to be labelled:

\begin{equation}
     \hat{P}_{I} = \repeatnn (P_{I}), N-1 \text{ times}
\end{equation}        

Then for the intent2slot we take the hidden state sequence of the remaining $N-1$ tokens, each of dimension $D_{E}$:

\begin{equation}
     H = (h_{2}, h_{3}, \ldots, h_{N})
\end{equation}  

We concatenate each element of $H$ with $P_{I}$ and feed it into a softmax classifier to get the slot probability for each word:

\begin{equation}
     S = (H \oplus \hat{P}_{I}) \cdot V_{S}^{T} + m_{S}
\end{equation} 

We have a matrix $S \in \mathbb{R}^{(N-1) \times D_{SP}}$. This represents $N-1$ vectors of size [$D_{E} + D_{IP}$] each being a BERT encoding concatenated with the intent probability. For each one there is a dense layer of size $D_{SP}$. Hence $V_{S} \in \mathbb{R}^{(N-1) \times D_{SP} \times (D_{E}+D_{IP})}$ is an array of $N-1$ matrices of weights and $m_{S} \in \mathbb{R}^{(N-1)\times D_{SP}}$ is an array of $n-1$ vectors of biases. The $N-1$ linear outputs feed to $N-1$ softmaxes, each with an argmax to get a predicted slot label for each element of the input sequence.

\begin{equation} \label{eq:S}
     S_{n}= \argmax(\softmax ({S^{n}} )), n \in \{2, \ldots , N\}
\end{equation} 

where $S^{n}$ is the $n$th slice of $S$ on the first axis.

\textbf{Slot2Intent model}
The slot2intent model aims to take the sequence label probability distributions as additional information in intent detection. This can be viewed as a dual of the intent2slot model. Similarly to the intent2slot model, for intent classification we first take the final hidden state sequence of the tokens excluding the first special token [CLS], each of dimension $D_{E}$:

\begin{equation}
     H = (h_{2}, h_{3}, \ldots, h_{N})
\end{equation} 

For each $h_{n} \in H$ we have a dense layer with $D_{SP}$ nodes. Then we use softmax to get the slot probabilities $P_{S_{n}}$:

\begin{equation}
     P_{S_{n}} = \softmax (h_{n} \cdot W_{S_{n}}^{T} + b_{S_{n}}), n \in \{2, \ldots , N\} 
\end{equation} 

Here each $W_{S} \in \mathbb{R}^{D_{SP} \times D_{E}}$ is a matrix of weights and each $b_{S} \in \mathbb{R}^{D_{SP}}$ is a vector of biases. Each $P_{S_{n}} \in \mathbb{R}^{D_{SP}}$ is thus a vector of slot label probability distributions for a token of the input sequence after [CLS].

In order to fuse the slot probabilities with the intent hidden state, we flatten the slot probabilities as:

\begin{equation}
     \hat{P}_{S} = \flatten (P_{S_{n}}),  n \in \{2, \ldots , N\} 
\end{equation} 

This is a vector of length $D_{SP} (N-1)$.

In the slot2intent model, we concatenate the final hidden state of the first special token [CLS], $h_{1}$, with $\hat{P}_{S}$ and feed it into a softmax classifier to get the intent classification:

\begin{equation} \label{eq:I}
     I = \argmax(\softmax((h_{1} \oplus \hat{P}_{S} ) \cdot V_{I}^{T}  + m_{I} )) 
\end{equation} 

Here $V_{I} \in \mathbb{R}^{D_{IP} \times (D_{E} + D_{SP} (N-1))}$ is a matrix of weights, and $m_{I} \in \mathbb{R}^{D_{IP}}$ is a vector of biases.

\textbf{Joint optimisation}
To jointly model intent classification and slot filling, the model is trained on the sum of slot labeling loss $L_{S}$ and of intent classification loss $L_{I}$ (i.e. $L = L_I+L_S$). The intent loss for a sample is:

\begin{equation}
L_{I} = -\sum_{i\in \mathbb{I}}{I(y_g=i) \log{y_i}}
\end{equation} 

\noindent where $I(y_g=i)$ is the indicator function that the gold intent for the sample is $i$, $y_i$ is the output from the softmax classifier in Equation (\ref{eq:I}) corresponding to intent $i$ for the sample. The slot loss for a sample is 
\begin{equation}
    L_{S} = -\sum^{N}_{n=2}\sum_{s\in \mathbb{S}}{I(y^n_g=s) \log{y^n_s}}
\end{equation} 

\noindent where $I(y^n_g=s)$ is the indicator function that token $n$ has gold slot label $s$ and $y^n_s$ is the output from the softmax classifier in Equation (\ref{eq:S}) corresponding to intent $s$ for token $n$ of the sample.

\section{Experiment}
In this section, we first evaluate our bi-directional joint intent classification and slot filling model against the current state-of-the-art models. Then we conduct an ablation study to reveal the contributions from the input embedding layer, NLU modeling layer and the bi-directional NLU layer of our model to the final performance.

\subsection{Datasets}
We ran our experiments on two widely-used and publicly available benchmark datasets. 
ATIS is a single-domain, single utterance dataset. It consists of flight information query utterances. The dataset consists of 4,478 training, 500 validation, and 893 testing samples respectively. There are 120 slot labels and 21 intents.
SNIPS\footnote{https://github.com/snipsco/nlu-benchmark/} is collected from the SNIPS voice assistant. In the SNIPS dataset, the training set contains 13,084 utterances, the validation set contains 700 utterances, and another 700 utterances are used as the test set. There are 72 slot labels and 7 intent types. The dataset is more complicated than the ATIS since it contains more than one domain and a large vocabulary. 





\subsection{Settings}
The BERT model we used is ``bert\_uncased\_L-12\_H-768\_A-12''. We used 0.1 dropout rate and initialize the weights with standard deviation of 0.02. We set the maximum sequence length (including special tokens) to 50 and train our model on the training set of the two datasets. For all experiments, the learning rate is 5e-5 and the optimizer is the same Adam optimizer used for BERT pre-training. We applied 10 epochs to ATIS and 20 to SNIPS. We applied the same hyper-parameters one of our baselines \cite{chen2019bert}.



Following the literature we measure intent accuracy, span based slot f1\footnote{The Python implementation of \textit{conlleval script} is used for F1-score: https://pypi.org/project/seqeval/}, and semantic accuracy wherein all slots and intent are correctly predicted for a positive result.

\begin{table}[t]
\caption{Natural language understanding (NLU) performance on ATIS and SNIPS-NLU datasets (\%).}
\begin{center}\scriptsize
\setlength{\tabcolsep}{5pt}
\renewcommand{\arraystretch}{1.3}
\newcommand{\CTPanel}[1]{%
\multicolumn{1}{>{\columncolor{cyan}}l}{#1}}
\setlength\fboxsep{2mm}
\colorbox[cmyk]{0.1,0,0,0}{%
\begin{tabular}{p{1.75cm}|ccc|ccc}
\hline\hline
\multicolumn{1}{c|}{\multirow{2}{*}{\textbf{Model}}} & \multicolumn{3}{c|}{\textbf{ATIS (10 epoch)}}                        & \multicolumn{3}{c}{\textbf{SNIPS (20 epoch)}}                       \\ \cline{2-7} 
\multicolumn{1}{c|}{}                                & \makecell{\textbf{Slot} \\ \textbf{(f1)}} & \makecell{\textbf{Intent} \\ \textbf{(acc)}} & \makecell{\textbf{Sent.} \\ \textbf{(acc)}} & \makecell{\textbf{Slot} \\ \textbf{(f1)}} & \makecell{\textbf{Intent} \\ \textbf{(acc)}} & \makecell{\textbf{Sent.} \\ \textbf{(acc)}} \\ \hline
\textbf{Joint Seq \cite{hakkani2016multi} }                                    & 94.3               & 92.6                  & 80.7                    & 87.3               & 96.9                  & 73.2                    \\ \hline
\textbf{Attn. \cite{liu2016attention}}                                 & 94.2               & 91.1                  & 78.9                    & 87.8               & 96.7                  & 74.1                    \\ \hline
\textbf{S-Gated F. \cite{goo2018slot}}                        & 94.8               & 93.6                  & 82.2                    & 88.8               & 97.0                  & 75.5                    \\ \hline
\textbf{S-Gated I. \cite{goo2018slot}}                      & 95.2               & 94.1                  & 82.6                    & 88.3               & 96.8                  & 74.6                    \\ \hline



\textbf{Capsule NN \cite{zhang2019capsule}}           & 95.2               & 95.0                  & 83.4                    & 91.8               & 97.3                  & 80.9                    \\ \hline\hline

\textbf{Joint Bert \cite{chen2019bert}}                                  & 96.1               & 97.5                  & 88.2                   & 97.0               & 98.6                  & 92.8                   \\ \hline

\textbf{Stack-Prop.\cite{qin2019stack}}                                   &96.1               & 97.5                  & 88.6                    & 97.0               & 99.0                  & 92.9                    \\ \hline

\textbf{Our Model}        & \textbf{96.3}               & \textbf{98.6}    & \textbf{88.6}                    & \textbf{97.2}               & \textbf{99.2}                  & 92.8                    \\ \hline\hline


\end{tabular}
}
\label{table:NLU performance}
\end{center}
\end{table} 

\subsection{Experiment 1: Joint model overall evaluation}

As shown in the Table~\ref{table:NLU performance} we achieved better performance on all tasks for both datasets. For ATIS, the intent accuracy and the sentence-level semantic accuracy are improved by 3.6\% and 5.2\% respectively compared to the highest from the previous non-BERT-based state-of-the-art model \cite{zhang2019capsule}. Comparatively, the slot F1 was improved by a smaller amount, being 1.1\%. Similarly for SNIPS, the sentence-level semantic accuracy was increased by 11.9\%, while the intent accuracy and slot filling F1 were improved by 1.9\% and 5.4\% respectively. We also found that our proposed model out-performs BERT-based joint models, including Joint BERT \cite{chen2019bert} and Stack-Propagation with BERT\cite{qin2019stack}. Note that Joint BERT is a pre-print and the performance is not stable based on different learning rate, and Stack-Propagation-BERT set the default epoch as 300. 

We found that the proposed bi-directional contextual contribution (slot2intent, intent2slot) is effective and outperformed baseline models. For the ATIS dataset, the intent label distribution is imbalanced, ranging from a highest frequency of 3309 samples for \textit{atis\_flight} to a lowest of 1 for \textit{atis\_cheapest}. This may negatively impact the overall intent classification performance. In addition, the number of slots in ATIS (120) is significantly larger than SNIPS (72) and those slots in ATIS only cover one domain which results in a high number of slots being mutually shared by multiple intents. For example, the \textit{atis\_flight} intent contains 66 slot types, 52 of which are repeated in other intents. The efficacy of the contextual clues provided by slot2intent flow in our proposed model might be undermined by such overlapping slots across different intents. Despite all this, utilizing those slots unique to specific intents still improves the intent classification performance, as is verified by the results. On the other hand, the slot filling F1 score improved by a relatively smaller amount. This may be related to the longer sequence length and larger number of slots to be labelled for a single intent in ATIS. However, the improvement can still validate the effectiveness of the intent2slot flow, and indicates that providing the intent-related contextual information may help with narrowing down from the large slot pool to much smaller candidate groups.

Compared to ATIS, SNIPS is more complex regarding its cross-domain topics. For example, intents \textit{BookRestaurant}, \textit{GetWeather} and \textit{PlayMusic} are from different domains. Due to this, the vocabulary size is much larger than in ATIS. The diversity of vocabulary in SNIPS for each intent provides useful information for intent classification and its intent distribution in the training dataset is well balanced. These factors point to why the intent classification accuracy in SNIPS is relatively higher than in ATIS. In addition to this, SNIPS has relatively small number of overlapping slots (only 11 slots are mutually shared with other intents, while ATIS has 79 such slots). With less overlapping slots, our slot2intent flow further provides discriminated contextual information regarding slots, resulting in a high accuracy of 99.2\% for intent classification. 

On the other hand, SNIPS has only 72 slots for the 7 intents. For a single intent, the maximum slot type count is 15 (only considering slots starting with `\textit{B-}') for \textit{BookRestaurant} while the minimum is only 2 for \textit{SearchCreativeWork}. With the intent2slot flow providing topic-related clues for slot filling, our proposed model improves the slot filling performance. Compared to the ATIS result, this improvement is a little bit lower. This could be caused by ambiguity of slot value in SNIPS. For example, the location entity (e.g. `Sydney') labelled as slot value for intent \textit{flight\_time} is quite obvious in ATIS because most of the location entities are labelled as CITY. However, the location entity labelled as a slot value for intent \textit{GetWeather} in SNIPS can be labelled as CITY, STATE, or COUNTRY. This might be confusing to the machine when appearing alone without using an explicit dictionary. 


\begin{figure}[t]
\includegraphics[width=0.47\textwidth]{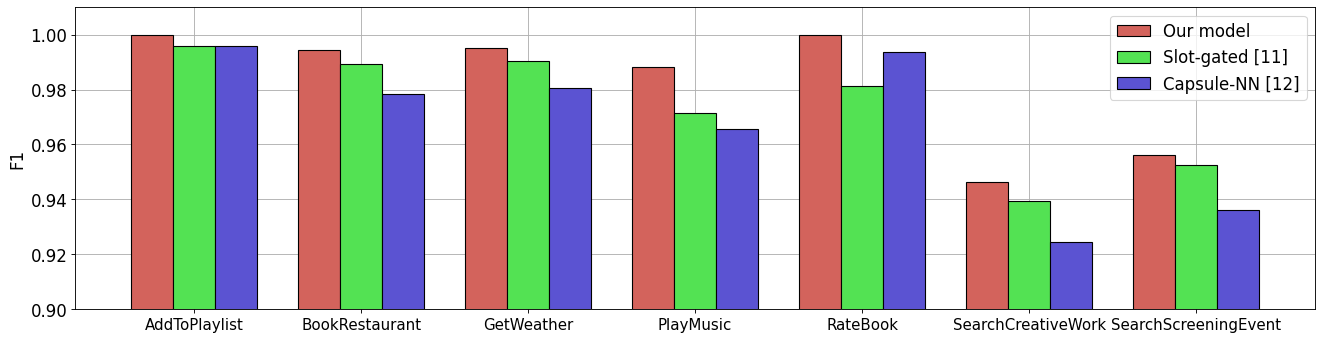}
\centering
\caption{Breakdown of intent detection performance of the proposed model against previous state-of-the-art models on SNIPS}
\label{fig:intent detect performance}
\end{figure}

\subsection{Experiment 2: Intent classification evaluation }

In order to further analyse the intent classification performance, we replicated the two most successful baseline models \cite{goo2018slot,zhang2019capsule} and trained them on the SNIPS dataset with 20 epochs, the same as for our proposed model. We measured the intent detection F1 score for each type of intent. As illustrated in Figure~\ref{fig:intent detect performance}, we observe a similar trend for intent F1 scores among all models. It is clear that \textit{SearchCreativeWork} and \textit{SearchScreeningEvent} always get significantly lower F1 than the other intents, which are uniformly high. We investigated the slot distributions and found that both the lower performing intents tend to have non-discriminative slots in the training datasets. For example, \textit{SearchCreativeWork} only has two slot types (starting with `\textit{B-}'), both of which also appear frequently in utterances of other intents. Similarly, \textit{SearchScreenEvent} has 3 out of its 7 slot types repeating in other intents. Further, one of the only two slot types for \textit{SearchCreativeWork} also appears as a slot type for \textit{SearchScreenEvent}. In addition, these two intents may  actually be about the same topic when the creative work is a movie. Thus confusion may occur. After we further looked at the classification results of the three models, we found a similar pattern: most of the \textit{SearchScreenEvent} were misclassified as \textit{SearchCreativeWork} and the exact case rates of \textit{SearchScreenEvent} misclassified as \textit{SearchCreativeWork} are 11 out of 14 for Capsule-NN, 5 out of 7 for Slot-gated and 5 out of 5 for our proposed model. This causes the negative impact on precision and recall for the two intents in joint models. 

Another notable point is that only for intent \textit{RateBook} does Capsule-NN outperform Slot-gated model. This might be related to the relatively short length and regular pattern of the \textit{RateBook} utterance. For instance, most of the utterances are similar to ``Give/Rate [1-9] point/star to Bookname''. This may lead to better slot filling performance. Additionally, \textit{RateBook} has a higher proportion of unique slots compared to most of the other intents. Given that, providing the slot-related clue, as Capsule-NN does, may contribute to better intent classification compared to other intents. This may also help explain the highest accuracy for \textit{RateBook} for our proposed model. It can be seen that our proposed model achieved better F1 for all intents, which validates the effectiveness of slot2intent as well as the mutual augmentation through bi-directional contextual flow.

\begin{table}[!h]
\caption{Ablation study results on ATIS}
\begin{center}\scriptsize
\setlength{\tabcolsep}{5pt}
\renewcommand{\arraystretch}{1.3}
\newcommand{\CTPanel}[1]{%
\multicolumn{1}{>{\columncolor{cyan}}l}{#1}}
\setlength\fboxsep{2mm}
\colorbox[cmyk]{0.1,0,0,0}{%
\begin{tabular}{c|c|c|c|c|c}
\hline\hline

\makecell{\textbf{NLU} \\ \textbf{Modeling}}  
& \textbf{Embedding} & 

\makecell{\textbf{Bidirectional} \\ \textbf{NLU}} & 

\makecell{\textbf{Slot} \\ \textbf{(f1)}} & \makecell{\textbf{Intent} \\ \textbf{(acc)}} & \makecell{\textbf{Sent.} \\ \textbf{(acc)}} \\ \hline

\multirow{4}{*}{LSTM} & GloVe              & X                      & 50.4                   & 70.8                     & 23.6                       \\ \cline{2-6} 
                      & GloVe              & \textbf{O}                      & \textbf{51.4}                   & \textbf{72.2}                     & \textbf{25.2}                       \\ \cline{2-6} 
                      & word2vec           & X                      & 40.4                   & 70.8                     & 13.8                       \\ \cline{2-6} 
                      & word2vec           & \textbf{O}                      & \textbf{44.5}                   & \textbf{72.0}                     & \textbf{16.3}                       \\ \hline
\multirow{4}{*}{GRU}  & GloVe              & X                      & 54.5                   & 70.8                     & 25.5                       \\ \cline{2-6} 
                      & GloVe              & \textbf{O}                      & \textbf{58.8}                   & \textbf{72.6}                     & \textbf{28.4}                       \\ \cline{2-6} 
                      & word2vec           & X                      & 45.0                   & 70.8                     & 18.1                       \\ \cline{2-6} 
                      & word2vec           & \textbf{O}                      & \textbf{54.1}                   & \textbf{71.9}                     & \textbf{25.5}                       \\ \hline
\multirow{2}{*}{BERT}
                  & BERT               & X                      & 94.8                   & 96.2                     & 86.1                      \\ \cline{2-6}
                  & BERT               & \textbf{O}                      & \textbf{96.3}                   & \textbf{98.6}                     & \textbf{88.6}                       \\ \hline\hline
\end{tabular}}

\label{tab:ablation testing ATIS}
\end{center}
\end{table}

\subsection{Experiment 3: Ablation study}
In order to investigate the effect of the input embedding layer, NLU modeling layer and the bidirectional NLU layer, we also report ablation study results in Table \ref{tab:ablation testing ATIS} on the ATIS dataset. 
It can be seen that we replace the input embedding layer from BERT to GloVe or word2vec, the NLU Modelling layer from BERT to Long Short-Term Memory (LSTM) or Gated Recurrent Unit (GRU). We also have each replaced input embedding layer and NLU Modelling Layer combinations with or without the bidirectional NLU layer. This aims to test the effectiveness of using mutual information in the model. 

The results show that the combination 'BERT NLU modeling + BERT embedding + Bidirectional NLU layer' outperformed all other combinations by a large margin on both datasets. By changing the BERT NLU modeling and BERT embedding to other combinations of GRU/LSTM with GloVe/word2vec, it can be seen that the performance regarding all the three metrics on both datasets decrease dramatically, without regard to the use of the bidirectional NLU layer. This indicates that BERT helps with the two tasks in terms of both word embedding and NLU modeling aspects. Besides, it can be observed through comparing the model pairs in column 'Bidirectional NLU' that applying the bidirectional NLU contributes to overall performance improvement for all models. This validates the effectiveness and value of the bidirectional mutual information sharing for the joint task. This improvement is especially significant for GloVe and word2vec models, neither of which capture contextual information to the depth that BERT does. 

\section{Conclusion}
In this paper, we propose a new and effective joint intent detection and slot filling model which integrates deep contextual embeddings and the transformer architecture. We further propose a Bi-directional NLU layer that contains two models: intent2slot for slot filling and slot2intent for intent classification. The model was applied to two real-world datasets and outperformed previous state-of-the-art results, using the same evaluation measurements: in intent detection, slot filling and semantic accuracy. The results indicate that our approach was more effective in capturing global interrelationships between the semantic components in the NLU tasks than existing joint models.

\bibliography{mybib}

\end{document}